\documentclass[letterpaper, 10 pt, conference]{ieeeconf}  

\IEEEoverridecommandlockouts                              

\usepackage{graphics} 
\usepackage{epsfig} 
\usepackage{amsmath}
\usepackage{amstext}
\usepackage{amssymb}
\usepackage{multirow}
\usepackage[hidelinks]{hyperref}
\usepackage{stmaryrd}
\usepackage{subfigure}

\title{\LARGE \bf
Matching Distance and Geometric Distribution Aided Learning Multiview Point Cloud Registration}

\author{Shiqi Li, Jihua Zhu, Yifan Xie, Naiwen Hu and Di Wang
\thanks{The authors are with School of Software Engineering, Xi'an Jiaotong University, Xi'an 710049, China
and Shaanxi Joint Key Laboratory for Artifact Intelligence, China
        {\tt\footnotesize (e-mail: lishiqi@stu.xjtu.edu.cn; zhujh@xjtu.edu.cn; xieyifan@stu.xjtu.edu.cn; naiwenhu@stu.xjtu.edu.cn; diwang@xjtu.edu.cn).}}%
}%

\begin{document}

\maketitle
\thispagestyle{empty}
\pagestyle{empty}

\begin{abstract}
    Multiview point cloud registration plays a crucial role in robotics, automation, and computer vision fields. This paper concentrates on pose graph construction and motion synchronization within multiview registration. Previous methods for pose graph construction often pruned fully connected graphs or constructed sparse graph using global feature aggregated from local descriptors, which may not consistently yield reliable results. To identify dependable pairs for pose graph construction, we design a network model that extracts information from the matching distance between point cloud pairs. For motion synchronization, we propose another neural network model to calculate the absolute pose in a data-driven manner, rather than optimizing inaccurate handcrafted loss functions. Our model takes into account geometric distribution information and employs a modified attention mechanism to facilitate flexible and reliable feature interaction. Experimental results on diverse indoor and outdoor datasets confirm the effectiveness and generalizability of our approach. The source code is available at \url{https://github.com/Shi-Qi-Li/MDGD}.
\end{abstract}


\def\degree{${}^{\circ}$}

\section{INTRODUCTION}

Multiview point cloud registration is a fundamental task in autonomous driving~\cite{vizzo2023kiss}, robotics~\cite{collet2009object}, and 3d computer vision~\cite{huang2021bundle}. When exploring unknown scenes with mobile robots or handheld scanners, a single scan typically captures only a local area, necessitating multiview registration to provide holistic scenario representation for downstream reconstruction~\cite{zeng20173dmatch}, semantic segmentation~\cite{landrieu2018large}, and mapping~\cite{wang2023hybridfusion}. Formally speaking, given $N$ unordered and partially overlapping point cloud frames, the multiview point cloud registration aims to recover the complete scene by locating a pose for each frame in a unified coordinate system. 

\begin{figure}
    \centering
    \includegraphics[width=\linewidth]{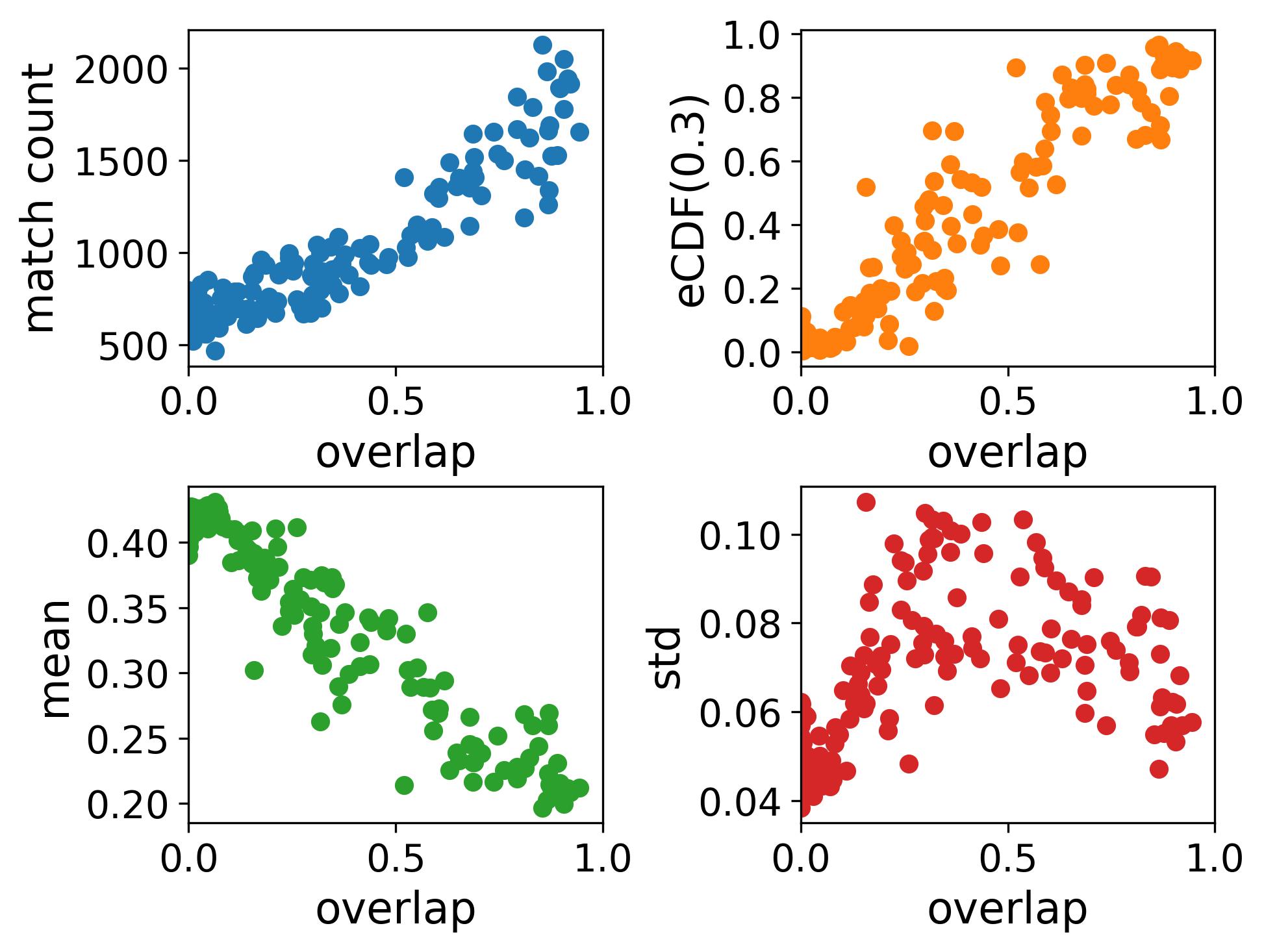}
    \caption{The number of matching and the empirical cumulative distribution function (eCDF) show a proportional relationship with overlap. Conversely, the average matching distance decreases as overlap increases. The standard deviation of the matching distance follows an approximate bell curve distribution.}
    \label{fig:dis}
\end{figure}

The multiview registration problem typically involves three subtasks: pairwise registration, pose graph construction, and motion synchronization. Pairwise registration, as a stepstone in multiview registration, seeks to determine the relative transformation between two given point cloud frames. Although recent learning and correspondence-based approaches~\cite{qin2022geometric,wang2022you} operate excellently in a variety of indoor and outdoor scenarios, low-overlapping and structureless scenes remain a formidable challenge for pairwise registration~\cite{huang2021predator}. Of course, subsequent multiview registration must carefully detect and disregard these poor pairwise outcomes; this relative transformation selection procedure is known as pose graph construction. Pruning from a fully connected graph~\cite{gojcic2020learning} and directly building a sparse graph~\cite{wang2023robust} are two prevalent paradigms for pose graph construction. The former utilizes some distance measures on the transformed point cloud to evaluate each predicted transformation, whereas the latter commonly employs a neural network model to estimate trustworthy pairs. As the final stage in multiview registration, motion synchronization further recognizes and eliminates outliers and noise from the input pose graph, resulting in final absolute poses~\cite{li2014improved}. Most studies use iteratively reweighted least squares (IRLS) based optimization techniques~\cite{chatterjee2017robust,huang2019learning} to handle this problem; nevertheless, weight initialization and cost function design remain unresolved problems. In addition to optimization-based approaches, some recent studies have attempted to solve the synchronization problem in a data-driven way~\cite{yew2021learning}. In this work, we concentrate on the pose graph construction and motion synchronization steps and propose a neural network model that mines the information from pairwise matching distance and geometric distribution clues to solve the multiview point cloud registration in a learning manner. 

The concept of mining information from descriptor matching distance derives from the common practice in learning and correspondence-based pairwise registration methods~\cite{huang2021predator,qin2022geometric,huang2022imfnet}. Most current registration methods typically extract descriptors on keypoints to establish correspondence set. During the training phase, metric learning loss is employed to bring descriptors associated with correctly matched keypoints together and push descriptors for unmatched points away. During registration, the correspondence is built upon the nearest match in the descriptor's feature space. Therefore, a natural intuition is that the distance between matched descriptors reflects the overlap information of a point cloud pair and the quality of calculated registration results. Specifically, a reliable pairwise registration tends to have a large correspondence set and a small distance between matched descriptors. Meanwhile, with an unreliable registration, the situation will be opposite. 
We match exhaustive pairs in the \emph{analysis-by-synthesis-apt2-luke} scene from 3DMatch~\cite{zeng20173dmatch} with YOHO~\cite{wang2022you} descriptor and collect the relevant matching distance data in each pair, some statistical information is illustrated in Fig. \ref{fig:dis}. 
Although the matching distance can reveal some pairwise information, it is still difficult to provide a specific formulation that explicitly quantifies the registration outcome. Therefore, we employ a tiny neural network to extract patterns from these distance statistical data and then recognize the reliable pairs for constructing pose graph. 

\begin{figure}
    \centering
    \includegraphics[width=\linewidth]{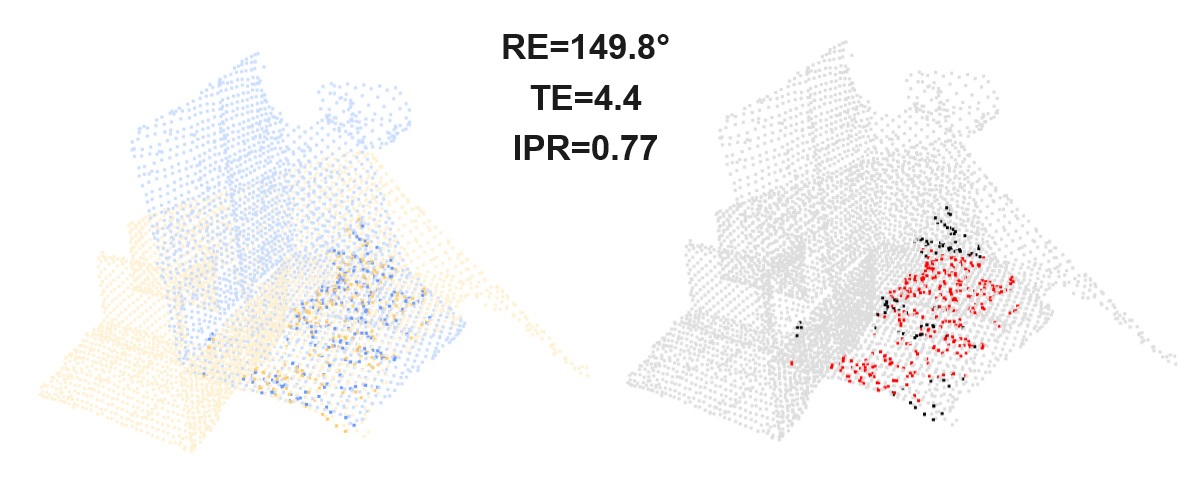}
    \caption{A failure registration case. Left: Point clouds registered by the predicted transformation. Darker colors represent keypoints whose associated correspondence confirms the predicted transformation. Right: Red points indicate keypoints lying in a fitting plane, while black points represent those do not lie in the plane. Better viewing with color and zooming in.}
    \label{fig:plane}
\end{figure}

Besides the matching distance, the geometric distribution can give additional information about pairwise registration. For pairwise registration, we anticipate that most correspondence may adapt to others while being constructed on similar geometric significant areas, such as corner points, between two frames. However, most point cloud feature extraction networks rely on a local aggregation design, which leads to points on the plane area having similar descriptors. Therefore, if the correspondences of a transformation are broadly located in a structureless area, the correctness of this pairwise registration is questionable. Fig. \ref{fig:plane} visualizes a failure example in which 77\% of the final consensus correspondences used to calculate the transformation are in a plane. Based on this concept, we collect data on the geometric distribution of a point cloud pair to further demonstrate its trustworthiness. Since geometric distribution-related information usually requires the predicted transformation, we only perform this operation on the constructed pose graph, and the resulting information is supplied into the motion synchronization as prior guidance.   

Motion synchronization is also a crucial component of multiview registration. Inspired by iterative optimization and recent advances in multiple rotation averaging areas~\cite{purkait2020neurora,li2022rago}, we propose an alternant feature update model to tackle motion synchronization in a data-driven fashion. Our model switches between absolute and relative feature updates and achieves a flexible data-dependent feature exchange fashion thanks to the introduction of confidence attention mechanism. Furthermore, we also incorporate the aforementioned matching distance and geometric distribution features to aid the optimization process. Extensive experiments on indoor and outdoor datasets demonstrate the accuracy and generalization ability of our proposed method.

In conclusion, our contribution can be summarized as: 
\begin{itemize}
    \item We propose a network that identifies reliable pairwise registrations for pose graph construction by extracting information from descriptor matching statistics.  
    \item We analyze the geometric distribution characteristics in pairwise registration and incorporate these insights into motion synchronization. 
    \item We design an effective and efficient data-driven motion synchronization model equipped with a modified attention mechanism.
\end{itemize}

\section{RELATED WORK}

\subsection{Pairwise Registration}

Pairwise point cloud registration, which serves as the foundation for multiview registration, can be broadly classified into two types. The first category includes correspondence-based approaches~\cite{qin2022geometric,huang2021predator,ao2023buffer}, which identify potential correspondences and obtain the associated transformation. Traditional techniques typically construct correspondences based on the distance between point clouds, however in the deep learning era, matching descriptors in feature space is a popular strategy. Additionally, a few notable works focus on identifying inlier correspondences from matching results~\cite{huang2022gmf,zhang20233d,jiang2023robust,wu2023sacf}. Regression-based approaches belong to the second category~\cite{huang2020feature,xu2022finet,wu2023rornet,wu2023correspondence,yuan2023egst}. These methods use neural networks to directly recover the transformation from point cloud pairs rather than generating correspondences. Regression-based approaches are capable of object-level registration, but they are less effective when applied to large-scale scenarios. As a result, our approach is based on correspondence-based pairwise registration techniques.   

\begin{figure*}[t]
    \centering
    \includegraphics[width=\linewidth]{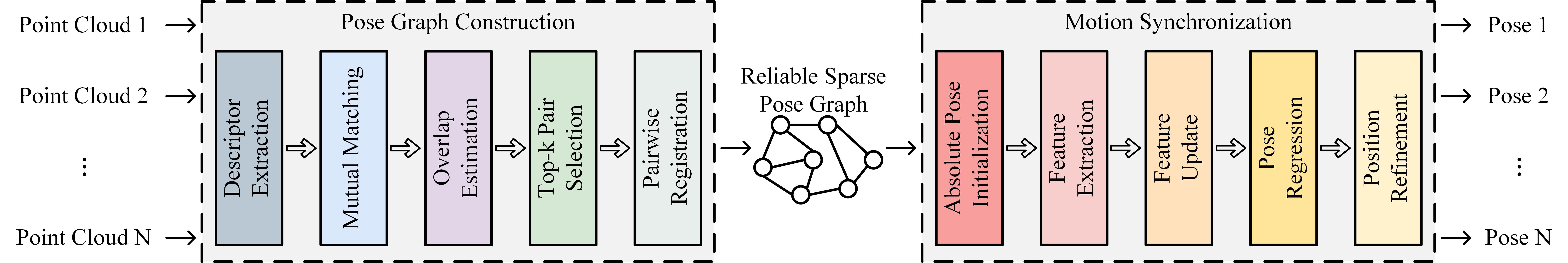}
    \caption{Pipeline for multiview point cloud registration with two parts: pose graph construction and motion synchronization.}
    \label{fig:main}
\end{figure*}

\subsection{Pose Graph Construction}

The pose graph construction procedure determines which pairwise registration result can contribute to downstream motion synchronization. The simplest option is a complete graph that contains all relative transformations between point cloud frames. However, such extensive pairwise registration is time-consuming and inevitably introduces enormous noise. Therefore, several outlier pruning techniques are proposed to clean the vanilla graph by detecting and removing false transformations~\cite{gojcic2020learning,yew2021learning}. As opposed to pruning from a fully connected graph, some other approaches use a neural network to construct a sparse but trustworthy pose graph directly~\cite{wang2023robust}. The network predicts overlap between each pair and selects only those with a high overlap for pairwise registration. Furthermore, some studies match point cloud frames in an incremental manner. However, these methods are usually hampered by the accumulation of errors during the growing process. 

\subsection{Motion Synchronization}

Motion synchronization is a specific group synchronization issue in $SE(3)$ space where the goal is to recover the absolute pose of each point cloud from the relative transformations in the constructed pose graph. The two-step (rotation-translation) procedure is commonly used in motion synchronization; the absolute orientations are first estimated by multiple rotation averaging, and the absolute locations are then derived. The majority of these methods employ the IRLS methodology~\cite{wang2023robust,zhu2021robust} to iteratively optimize a robust cost function. As an alternative to the two-step method, some solutions use spectral composition~\cite{arrigoni2016spectral} or data-driven~\cite{yew2021learning} methods to achieve $SE(3)$ synchronization directly. In this study, we use a neural network to predict absolute poses and refine the regressed positions with the least squares method.

\section{Method}

The pipeline of our method is illustrated in Fig. \ref{fig:main}. First, we use an off-the-shelf correspondence-based pairwise point cloud registration method to extract descriptors from each input point cloud frame and select reliable pairs using the matching-aided overlap estimation module. Subsequently, we calculate relative transformations for these chosen pairs using a robust estimator. Finally, the pairwise registration results are fed into the motion synchronization module, which calculates the absolute pose of each point cloud. 

\begin{figure*}
    \centering
    \includegraphics[width=\textwidth]{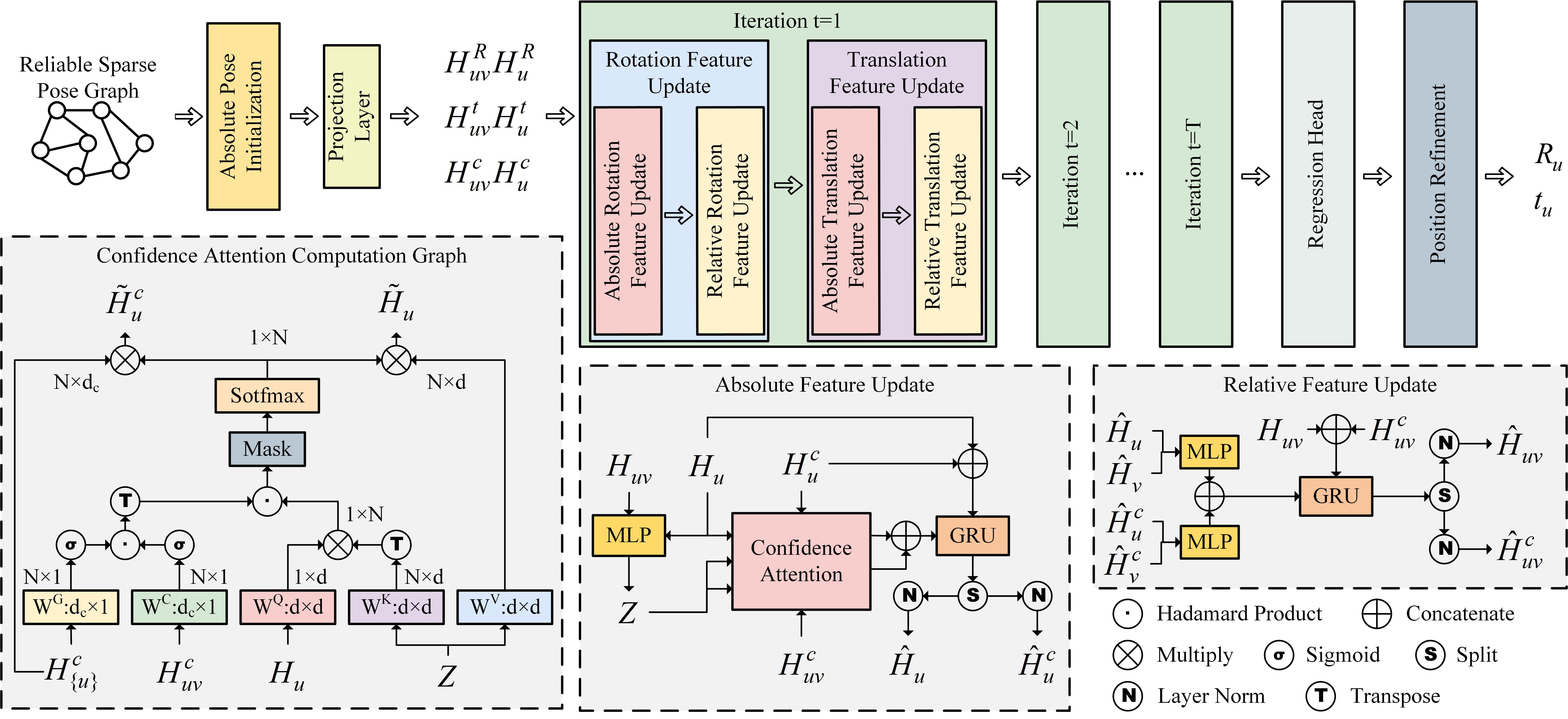}
    \caption{Motion synchronization module architecture. The pose graph is first initialized and projected to feature space. Then the alternate rotation and translation feature update is applied, both including confidence attention-based absolute update and relative update. The tilde notation indicates features after the confidence attention module, while the hat notation indicates features after the update.}
    \label{fig:network}
\end{figure*}

\subsection{Reliable Pose Graph Construction}
Selecting appropriate relative pairs is critical for multiview registration, both in terms of accuracy and efficiency. Since it is difficult to forecast the quality of a pairwise registration directly, the overlap ratio between two point cloud frames is typically employed as an alternative because pairs with high overlap are easier to handle. In this study, we use the sparse graph paradigm in~\cite{wang2023robust}, which only conducts pairwise registration on those pairs with high overlap ratio.

\subsubsection{Descriptor Extraction and Correspondence Establish}
Before feature matching and registration between any two point cloud frames, we need to first extract a set of descriptors for each point cloud $\Tilde{P}$. Because of the massive magnitude of point cloud data, extracting descriptors for each point is usually unaffordable. Therefore, we downsample a smaller keypoint set $P$ from the raw point cloud $\Tilde{P}$ and use the feature set $F$ associated with the keypoints $P$ to characterize the entire point cloud frame. 

Given descriptor sets $F_u$ and $F_v$, the correspondence $C_{uv}$ between frames $u$ and $v$ can be established via mutual nearest matching in the descriptor's feature space,
\begin{equation}
    \begin{aligned}
    C_{uv}=\{(i,j)|\text{NN}(f_i^u,F_v)=j\wedge\text{NN}(f_j^v,F_u)=i, \\
    \forall f_i^u\in F_u, \forall f_j^v\in F_v\},
    \end{aligned}
\end{equation}
where $\text{NN}(\cdot,\cdot)$ is nearest neighbour search.

\subsubsection{Overlap Estimation}
For each correspondence $C_{uv}$, the matching distance $D_{uv}$ is computed and converted to the input of our overlap prediction module. The matching distance is simply defined as the Euclidean distance between matched descriptors. Then we collect a variety of statistical data to holistically describe the quality of pairwise matching. Specifically, we count the mean, median, and standard deviation of $D_{uv}$. To better depict the distribution of matching distance, we further count the cumulative distribution function at some predetermined thresholds $\epsilon$. Finally, the cardinal number of set $C_{uv}$ is also used as a clue. Using aforementioned statistical data, we predict the overlap ratio $o_{uv}$ between point cloud frame $u$ and $v$ by 
\begin{equation}
    o_{uv}=\phi(|C_{uv}|, \Bar{D}_{uv}, \Tilde{D}_{uv}, \sigma_{D_{uv}}, Y_{D_{uv}}(\epsilon)),
    \label{eq:overlap}
\end{equation}
where $\Bar{D}_{uv}$, $\Tilde{D}_{uv}$, and $\sigma_{D_{uv}}$ are mean, median and standard deviation of $D_{uv}$, $Y_{D_{uv}}(\epsilon)$ denotes the eCDF of $D_{uv}$ at $\epsilon$. $\phi$ is a network consisting of an MLP projection layer, three stacked inverted residual blocks, and an MLP regression head. The inverted residual block expands the input feature dimension by a factor of 4, then compresses it back to the original dimension, utilizing a residual connection to ensure training stability. 

\subsubsection{Pairwise Selection}
After predicting the overlap, we create the pose graph $\mathcal{G=\{E,V\}}$ by performing pairwise registration between each frame and the other $k$ frames with the highest overlap ratio,
\begin{equation}
    \mathcal{E}=\{(u,v:\mathop{\text{arg-topk}}\limits_{v\neq u} o_{uv})|1\leq u\leq N\}.
    \label{eq:graph}
\end{equation}

\subsection{Pairwise Registration}
For each selected reliable pair, we use the existing robust estimator to calculate the relative transformation. The relative transformation $T_{uv}$ is estimated using RANSAC~\cite{fischler1981random} based on correspondence $C_{uv}$ established by descriptor matching. Additionally, Procrustes~\cite{schonemann1966generalized} refinement, which focuses on minimizing the distance between the overlap area, is also applied to improve registration accuracy. 

\subsection{Motion Synchronization}
In this section, we aim to recover the absolute orientation and position of each point cloud frame from the sparse but more trustworthy pose graph that we have built. 

\subsubsection{Initial Feature Extraction}
For each selected pairwise transformation $T_{uv}$, the rotation matrix $R_{uv}$ and translation vector $t_{uv}$ are separately projected to $H^R_{uv}$ and $H^t_{uv}$ by two multilayer perceptrons (MLP).  In addition, to demonstrate the reliability of each relative registration, we build a confidence feature based on the predicted overlap ratio $o_{uv}$, inlier correspondence ratio (ICR), and inlier plane ratio (IPR). Here, the ICR is defined as the ratio of correspondence conforming with the estimated transformation,
\begin{equation}
    \text{ICR}_{uv}=\sum_{(i,j)\in C_{uv}}\llbracket \Vert p^u_i-T_{uv}p^v_j\Vert_2 < \tau \rrbracket/|F_{u}|,
    \label{eq:icr}
\end{equation}
where $\llbracket\cdot\rrbracket$ is Iverson bracket which takes 1 only if the statement is true, $\tau$ is a distance threshold.

For IPR, we transform the points associated with the inlier correspondence defined in Eq. \ref{eq:icr} to the same coordinate and merge them to get a point cloud $P_{uv}$. Then we segment the largest plane $I_{uv}$ in this combined point cloud, and the IPR is defined as the ratio of points located in that plane,
\begin{equation}
    \text{IPR}_{uv}=\sum_{p\in P_{uv}}\llbracket d(p,I_{uv}) < \kappa \rrbracket/|P_{uv}|,
    \label{eq:ipr}
\end{equation}
where $d(p,I_{uv})$ is the distance between point $p$ and plane $I_{uv}$, $\kappa$ is a distance threshold. 

Finally, we stack $o_{uv}$, $\text{ICR}_{uv}$, and $\text{IPR}_{uv}$ data and feed them into another MLP to generate confidence feature $H^C_{uv}$.

Besides the relative features, we assign an initialization to each point cloud frame that can help the subsequent optimization process. Formally, we build a maximum-spanning tree on the pose graph. The edge priority $s_{uv}$ is determined as the product of predicted overlap $o_{uv}$ and $\text{ICR}_{uv}$.

After building the spanning tree, the root with an identity matrix pose is assigned to the tree's centroid. By searching from the root, we can obtain the initial pose of all frames. During the search process, we additionally gather the hop count and cumulative product of $s_{uv}$ along the path from each node to the root, the two attributes are further utilized to represent the trustworthiness of each absolute pose. To obtain absolute rotation, translation and confidence features, we use three more MLPs in the same way as we did while creating relative features. 

\subsubsection{Iterative Optimization Network}

Upon obtaining the initial features, the iterative optimization network is utilized to mine patterns from the redundant pose graph using information that flows between absolute and relative features. As illustrated in the Fig. \ref{fig:network}, the network alternately updates rotation and translation features, with both updating processes consisting of absolute and relative feature revisions. 

Given the rotation feature update procedure as an example, we update the absolute feature first, followed by the relative feature. Our absolute feature update design is inspired by the idea of solving multiple rotation averaging by iteratively performing several single rotation averaging~\cite{hartley2011l1}; here, we emulate this approach in feature space rather than $SO(3)$ space. Formally, for node $u$, we collect its neighbor node features $\{H^R_{v}|v\in\mathcal{N}_{u}\}$ and the related edge features $\{H^R_{uv}|v\in\mathcal{N}_u\}$ that connect them, a shared MLP is used to mix them and generate intermediate features $Z$. Then we use a modified confidence attention mechanism to generate new absolute rotation and confidence features. Compared to vanilla attention, our solution incorporates the confidence features, resulting in more flexible data and confidence-dependent feature updates, which guides the model's emphasis on reliable transformations. The architecture of our confidence attention is illustrated in Fig. \ref{fig:network}. 

After integrating the new features, we use a Gated Recurrent Unit (GRU)~\cite{cho2014learning} module to update the prior features rather than replacing them directly. The rotation and confidence features are concatenated so that the GRU can update them simultaneously. Furthermore, we also use two individual Layernorm~\cite{ba2016layer} operations to improve the stability. 

Relative feature updates come after absolute feature refreshes. Compared to the complex absolute feature creation, the new relative rotation or confidence features can be simply obtained by fusing features from nodes $u$ and $v$. The similar GRU and Layernorm designs are also used in this stage. 

Since the translation is affected by the rotation, the translation feature is renewed after the rotation feature. In the translation feature update, the absolute features are updated first, followed by the relative features. This is consistent with the rotation part.   

Our iterative optimization network repeats the aforementioned alternative process $T$ rounds to ensure that all features converge to a stable and optimal state. It is worth noting that different iteration rounds share the same weights. 

\subsubsection{Pose Generation}

We regress the absolute orientations $R_u$ and positions $t_u$ based on the corresponding features $H^R_u$ and $H^t_u$. Considering the continuity of rotation representation, we use the 6D representation proposed in~\cite{zhou2019continuity}. Specifically, we use an MLP to regress a 6D vector from the absolute rotation feature and convert it to a rotation matrix via Gram-Schmidt orthogonalization. Another MLP simply regresses the 3D position.    

In addition, we set an auxiliary task that predicts relative rotations $R_{uv}$ and translations $t_{uv}$ from features $H^R_{uv}$ and $H^t_{uv}$.

Furthermore, another MLP is used to regress weight factors $w_{uv}$ based on the relative confidence feature $H^c_{uv}$.  

During the training phase, we execute all these regressions in each iteration round, but in the test phase, we only regress once at the final iteration.  

\subsubsection{Position Refinement}

Since neural networks are notoriously bad at regressing precise quantities, we use the least square method to refine the predicted coarse position estimation of each point cloud frame~\cite{gojcic2020learning,wang2023robust,huang2019learning}.  

We construct three zero matrices, $\mathbf{P}\in\mathbb{R}^{3|\mathcal{E}|\times 3|\mathcal{E}|}$, $\mathbf{B}\in\mathbb{R}^{3|\mathcal{E}|\times 3N}$, and $\mathbf{L}\in\mathbb{R}^{3|\mathcal{E}|\times 1}$, where $\mathbf{P}$ and $\mathbf{B}$ are block matrices composed of $3\times 3$ blocks, and $\mathbf{L}$ is composed of $3\times1$ blocks. For $e$-th edge $(u,v)$ in $\mathcal{E}$, the $e$-th block on the diagonal of $\mathbf{P}$ is set to $w_{uv}\mathbf{I}_3$, the $(e,u)$ and $(e,v)$ block of $\mathbf{B}$ is filled by $R_{u}^\top$ and $-R_v^\top$, $e$-th block of $\mathbf{L}$ is filled by $R^\top_ut_{uv}-R^\top_ut^{croase}_u+R_v^\top t^{croase}_v$. Finally, we refine the $t^{corase}$ by
\begin{equation}
    t=t^{corase}+(\mathbf{B}^\top\mathbf{P}\mathbf{B})^{-1}\mathbf{B}^\top\mathbf{P}\mathbf{L}.
\end{equation}

\subsection{Loss Function}
\subsubsection{Overlap Loss} 
To train the overlap estimation module, we apply the smooth $L1$ loss~\cite{girshick2015fast} between ground truth and predicted overlap ratio,
\begin{equation}
    \mathcal{L}_{o}=\frac{1}{N^2}\sum_{u=1}^N\sum_{v=1}^N\text{smooth}_{L1}(o_{uv}-o_{uv}^{gt}).
\end{equation}

\subsubsection{Motion Loss}

We use the ground truth relative transformations to train our motion synchronization network. The rotation loss is defined as:  
\begin{equation}
    \begin{aligned}
        \mathcal{L}_{rot}=\frac{1}{N^2}\sum_{t=1}^{T}\sum_{u=1}^N\sum_{v=1}^N\gamma^{T-t}|R_{u}R_{v}^\top-R_{uv}^{gt}| \\
        +\frac{1}{|\mathcal{E}|}\sum_{t=1}^{T}\sum_{(u,v)\in\mathcal{E}}\gamma^{T-t}|R_{uv}-R_{uv}^{gt}|,
    \end{aligned}
\end{equation}
where $\gamma$ is a discounting factor.
The translation loss is similar to the rotation loss but has an additional term to supervise the results after the least square refinement,
\begin{equation}
   \begin{aligned}
       \mathcal{L}_{trans}=\frac{1}{N^2}\sum_{t=1}^{T}\sum_{u=1}^N\sum_{v=1}^N\gamma^{T-t}|t_u-R_{u}R_{v}^\top t_{v}-t_{uv}^{gt}| \\
       +\frac{1}{N^2}\sum_{t=1}^{T}\sum_{u=1}^N\sum_{v=1}^N\gamma^{T-t}|t^{croase}_u-R_{u}R_{v}^\top t^{croase}_{v}-t_{uv}^{gt}| \\
        +\frac{1}{|\mathcal{E}|}\sum_{t=1}^{T}\sum_{(u,v)\in\mathcal{E}}\gamma^{T-t}|t_{uv}-t_{uv}^{gt}|.
   \end{aligned}
\end{equation}

Finally, the total motion loss is a weighted sum of rotation loss and translation loss,
\begin{equation}
\mathcal{L}_{m}=\mathcal{L}_{rot}+\beta \mathcal{L}_{trans}, 
\end{equation}
where $\beta$ is a weight factor.

\section{Experiments}

\begin{table}[t]
\caption{Registration Recall on 3DMatch and ETH. Results are cited from~\cite{wang2023robust}, bold denotes best.}
\label{tab:3dmatch}
\centering
\resizebox{\linewidth}{!}{
\begin{tabular}{c|ccc|ccc|ccc}
\hline
\multirow{2}{*}{Method} & \multicolumn{3}{c|}{3DMatch} & \multicolumn{3}{c|}{3DLoMatch} & \multicolumn{3}{c}{ETH} \\
                        & Full   & Pruned   & Sparse   & Full    & Pruned   & Sparse & Full   & Pruned   & Sparse   \\ 
\hline
EIGSE3~\cite{arrigoni2016spectral}  & 23.2 & 40.1 & 60.4 & 6.6  & 26.5 & 44.6 & 60.9 & 96.3 & - \\
L1-IRLS~\cite{chatterjee2017robust} & 52.2 & 68.6 & -    & 32.2 & 49.0 & -    & 77.2 & 90.2 & - \\
RotAvg~\cite{chatterjee2017robust}  & 61.8 & 77.2 & 81.7 & 44.1 & 60.3 & 63.9 & 85.4 & 96.6 & - \\
LITS~\cite{yew2021learning}         & 77.0 & 80.8 & 84.6 & 59.0 & 65.2 & -    & 34.8 & 48.4 & - \\
HARA~\cite{lee2022hara}             & 83.1 & 83.8 & -    & 68.7 & 71.9 & -    & 85.4 & 96.0 & - \\
SGHR~\cite{wang2023robust}          & 93.2 & 95.2 & \textbf{96.2} & 76.8 & \textbf{82.3} & 81.6 & 98.8 & 97.2 & 99.1 \\
Ours                                & 95.0 & -    & \textbf{96.2} & 80.8 & -  & 81.2 & 87.3 & - & \textbf{99.3}\\
\hline
\end{tabular}
}
\end{table}

\begin{table*}[t]
    \caption{Registration results on Scannet. Results are cited from~\cite{wang2023robust}, bold denotes best.}
    \label{tab:scannet}
    \centering
    \begin{tabular}{c|c|cccccc|cccccc}
    \hline
    \multirow{2}{*}{pose} & \multirow{2}{*}{method} & \multicolumn{6}{c|}{Rotation Error} & \multicolumn{6}{c}{Translation Error ($m$)}            \\
                          &                         & 3\degree & 5\degree & 10\degree & 30\degree & 45\degree & Mean/Med & 0.05 & 0.1 & 0.25 & 0.5 & 0.75 & Mean/Med \\
    \hline
    \multirow{8}{*}{Full} & LMVR~\cite{gojcic2020learning}      & 48.3 & 53.6 & 58.9 & 63.2 & 64.0 & 48.1\degree/33.7\degree & 34.5 & 49.1 & 58.5 & 61.6 & 63.9 & 0.83/0.55 \\
                          & EIGSE3~\cite{arrigoni2016spectral}  & 19.7 & 24.4 & 32.3 & 49.3 & 56.9 & 53.6\degree/48.0\degree & 11.2 & 19.7 & 30.5 & 45.7 & 56.7 & 1.03/0.94 \\
                          & L1-IRLS~\cite{chatterjee2017robust} & 38.1 & 44.2 & 48.8 & 55.7 & 56.5 & 53.9\degree/47.1\degree & 18.5 & 30.4 & 40.7 & 47.8 & 54.4 & 1.14/1.07 \\
                          & RotAvg~\cite{chatterjee2017robust}  & 44.1 & 49.8 & 52.8 & 56.5 & 57.3 & 53.1\degree/44.0\degree & 28.2 & 40.8 & 48.6 & 51.9 & 56.1 & 1.13/1.05 \\
                          & LITS~\cite{yew2021learning}         & 52.8 & 67.1 & 74.9 & 77.9 & 79.5 & 26.8\degree/27.9\degree & 29.4 & 51.1 & 68.9 & 75.0 & 77.0 & 0.68/0.66 \\
                          & HARA~\cite{lee2022hara}             & 54.9 & 64.3 & 71.3 & 74.1 & 74.2 & 32.1\degree/29.2\degree & 35.8 & 54.4 & 66.3 & 69.7 & 72.9 & 0.87/0.75 \\
                          & SGHR~\cite{wang2023robust}          & 57.2 & 68.5 & 75.1 & 78.1 & 78.8 & 26.4\degree/19.5\degree & 39.4 & 61.5 & 72.0 & 75.2 & 77.6 & 0.70/0.59 \\
                          & Ours                    & 54.7 & 71.4 & 83.4 & 88.2 & 88.6 & 17.6\degree/19.1\degree & 38.7 & 62.8 & \textbf{77.8} & 82.6 & \textbf{85.2} & 0.42/0.35 \\
    \hline
    \multirow{6}{*}{Pruned} & EIGSE3~\cite{arrigoni2016spectral} & 40.8 & 46.3 & 51.9 & 61.2 & 65.7 & 40.6\degree/37.1\degree & 23.9 & 38.5 & 51.0 & 59.3 & 66.1 & 0.88/0.84 \\
                          & L1-IRLS~\cite{chatterjee2017robust}  & 46.3 & 54.2 & 61.6 & 64.3 & 66.8 & 41.8\degree/34.0\degree & 24.1 & 38.5 & 48.3 & 55.6 & 60.9 & 1.05/1.01 \\
                          & RotAvg~\cite{chatterjee2017robust}   & 50.2 & 60.1 & 65.3 & 66.8 & 68.8 & 38.5\degree/31.6\degree & 31.8 & 49.0 & 58.8 & 63.3 & 65.6 & 0.96/0.83 \\
                          & LITS~\cite{yew2021learning}          & 54.3 & 69.4 & 75.6 & 78.5 & 80.3 & 24.9\degree/19.9\degree & 31.4 & 54.4 & 72.3 & 76.7 & 79.6 & 0.65/0.56 \\
                          & HARA~\cite{lee2022hara}              & 55.7 & 63.7 & 69.0 & 70.8 & 72.1 & 34.7\degree/31.3\degree & 35.2 & 53.6 & 65.4 & 68.6 & 71.7 & 0.86/0.71 \\
                          & SGHR~\cite{wang2023robust}           & \textbf{59.4} & 71.9 & 80.0 & 82.1 & 82.6 & 21.7\degree/19.1\degree & \textbf{39.9} & 63.0 & 74.3 & 77.6 & 80.2 & 0.64/0.47 \\
    \hline
    \multirow{2}{*}{Sparse} & SGHR~\cite{wang2023robust}         & 59.1 & \textbf{73.1} & 80.8 & 82.5 & 83.0 & 21.7\degree/\textbf{19.0}\degree & \textbf{39.9} & \textbf{64.1} & 76.7 & 79.0 & 81.9 & 0.56/0.49 \\
                          & Ours                    & 56.1 & 71.8 & \textbf{83.5} & \textbf{88.5} & \textbf{88.8} & \textbf{17.4}\degree/\textbf{19.0}\degree & 38.2 & 61.2 & 77.5 & \textbf{82.7} & 84.9 & \textbf{0.37}/\textbf{0.31} \\
    \hline
    \end{tabular}
    
\end{table*}

\subsection{Dataset}

We conduct experiments on three different datasets: 3D(Lo)Match~\cite{zeng20173dmatch,huang2021predator}, ScanNet~\cite{dai2017scannet}, and ETH~\cite{pomerleau2012challenging}.  

3DMatch is an indoor dataset that is widely used in registration tasks. It contains 46 training scenes, 8 validation scenes, and 8 testing scenes, respectively. We follow the common practice that evaluates both the 3DMatch and 3DLoMatch test split. The overlap of point cloud pairs in 3DMatch is larger than $30\%$, while in 3DLoMatch is between $10\%$ and $30\%$.  

ScanNet is an RGB-D dataset collected from more than 1500 indoor scenes. We follow the practice of previous works~\cite{gojcic2020learning,wang2023robust} to test 32 scenes containing 960 scans and 13920 pairs in total.  

ETH is an outdoor point cloud dataset containing 4 scenes, each scene has about 33 scans and 713 pairs are officially selected for evaluation. 

\subsection{Baseline and Protocol} 

We compare our method with EIGSE3~\cite{arrigoni2016spectral}, L1-IRLS~\cite{chatterjee2017robust}, RotAvg~\cite{chatterjee2017robust}, LMVR~\cite{gojcic2020learning}, LITS~\cite{yew2021learning}, HARA~\cite{lee2022hara}, and SGHR~\cite{wang2023robust}. We use YOHO~\cite{wang2022you} to provide pairwise registration for all methods except LMVR which is an end-to-end method that achieves both pairwise and multiview registration in one network.  

For pose graph type, ``Full" means exhaustively register all scan pairs. ``Pruned" means only retaining scan pairs whose median point distance in the registered overlapping region is less than 0.05m (0.15m for ETH) from the fully connected graph~\cite{gojcic2020learning}. ``Sparse" means just selecting the pairs based on the estimated overlap score.  

Similar to the~\cite{wang2023robust}, our model is only trained on the training split of 3DMatch and tested on all datasets.

\subsection{Metrics}

Following~\cite{gojcic2020learning,wang2023robust}, we evaluate the multiview registration by relative transformation calculated from the recovered absolute pose.   

For 3D(Lo)Match and ETH datsets, we report Registration Recall (RR) metric. The RR reports the fraction of point cloud pairs whose transformation error is smaller than a certain threshold (0.2m for 3D(Lo)Match and 0.5m for ETH).  

For ScanNet, we report the mean, median, and eCDF of the rotation error $re$ and translation error $te$. The $re$ and $te$ are defined as:   

\begin{equation}
    re=\arccos\left(\frac{tr(R^{pred\top}R^{gt})-1}{2}\right), te=\Vert t^{pred}-t^{gt}\Vert_2.
\end{equation}

\subsection{Implementation Details}
Since our model is data-driven, we synthesize a dataset based on 3DMatch for training. For each scene in the 3DMatch training split, we generate random transformations by sampling three Euler angle rotations in the range [-180\degree, 180\degree] and translations within [-6, 6] on each axis for every frame. We repeat the process 150 times to obtain the pose graph training dataset.   

For the overlap estimation module, we train 1 epoch on the synthetic dataset. Then we freeze the weight of overlap module and train the motion synchronization module for 15 epochs. We use the AdamW optimizer, the initial learning rates of overlap module and motion module are 0.01 and 0.001, respectively. The weight decay is set to 0.0001 and the cosine learning rate schedule is also applied. 

For the eCDF threshold $\epsilon$ in Eq. \ref{eq:overlap}, we use four different values: 0.25, 0.3, 0.35, and 0.4. We set $k$ in Eq. \ref{eq:graph} to 10 for indoor datasets and 6 for outdoor dataset. The distance threshold $\tau$ and $\kappa$ in Eq. \ref{eq:icr} and Eq. \ref{eq:ipr} are set to 0.07 and 0.01, respectively. The iteration rounds $T$ of the motion synchronization module are set to 4, and in each rotation/translation feature update process, we conduct two absolute updates and one relative update. The discounting factor $\gamma$ and weight factor $\beta$ in loss functions are set to 0.8 and 0.2, respectively. 

\begin{figure*}[h]
    \centering
    \subfigure[Raw]{
    \begin{minipage}[b]{0.23\linewidth}
    \includegraphics[width=1\linewidth]{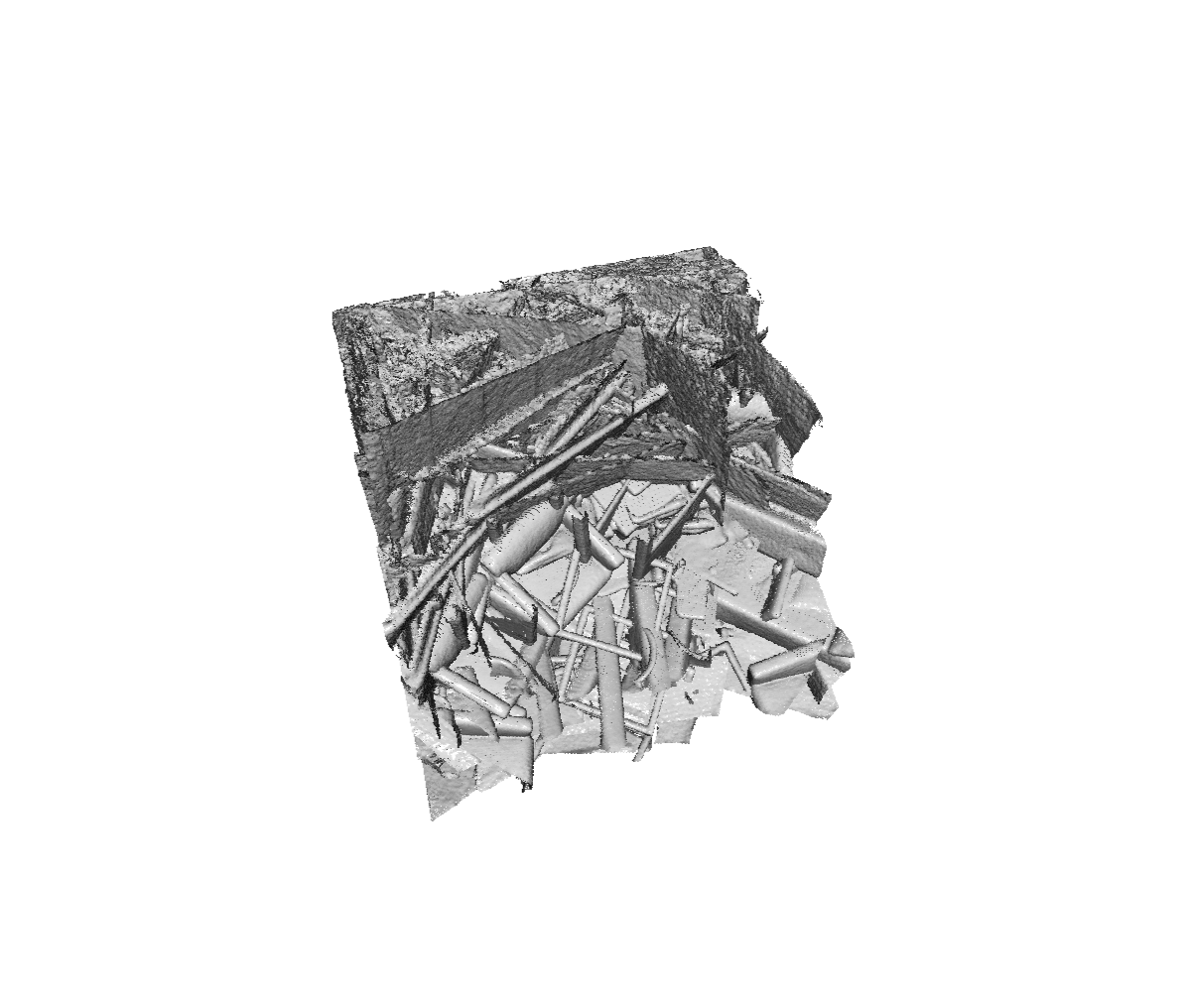}
    \includegraphics[width=1\linewidth]{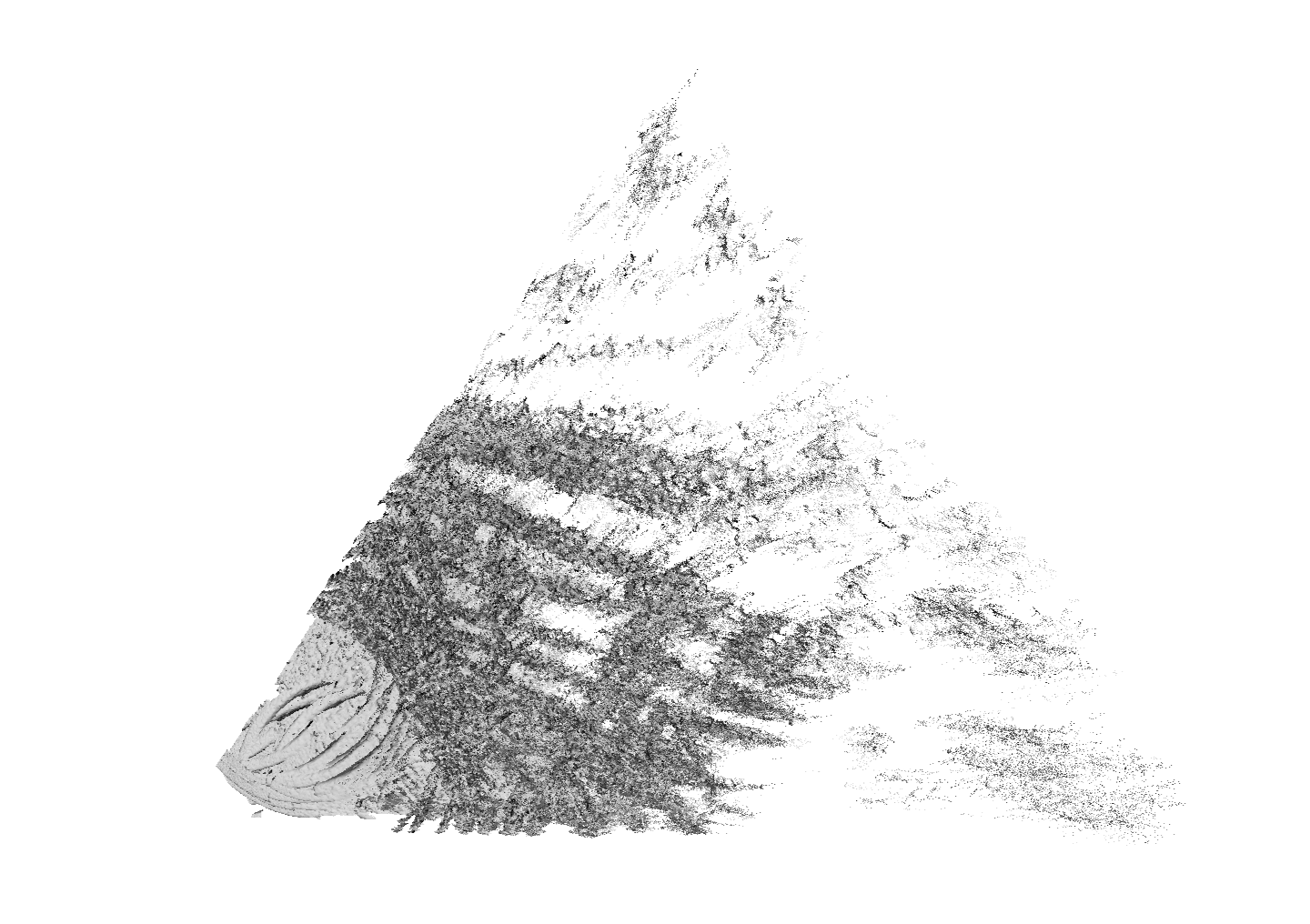}
    \end{minipage}}
    \subfigure[SGHR]{
    \begin{minipage}[b]{0.23\linewidth}
    \includegraphics[width=1\linewidth]{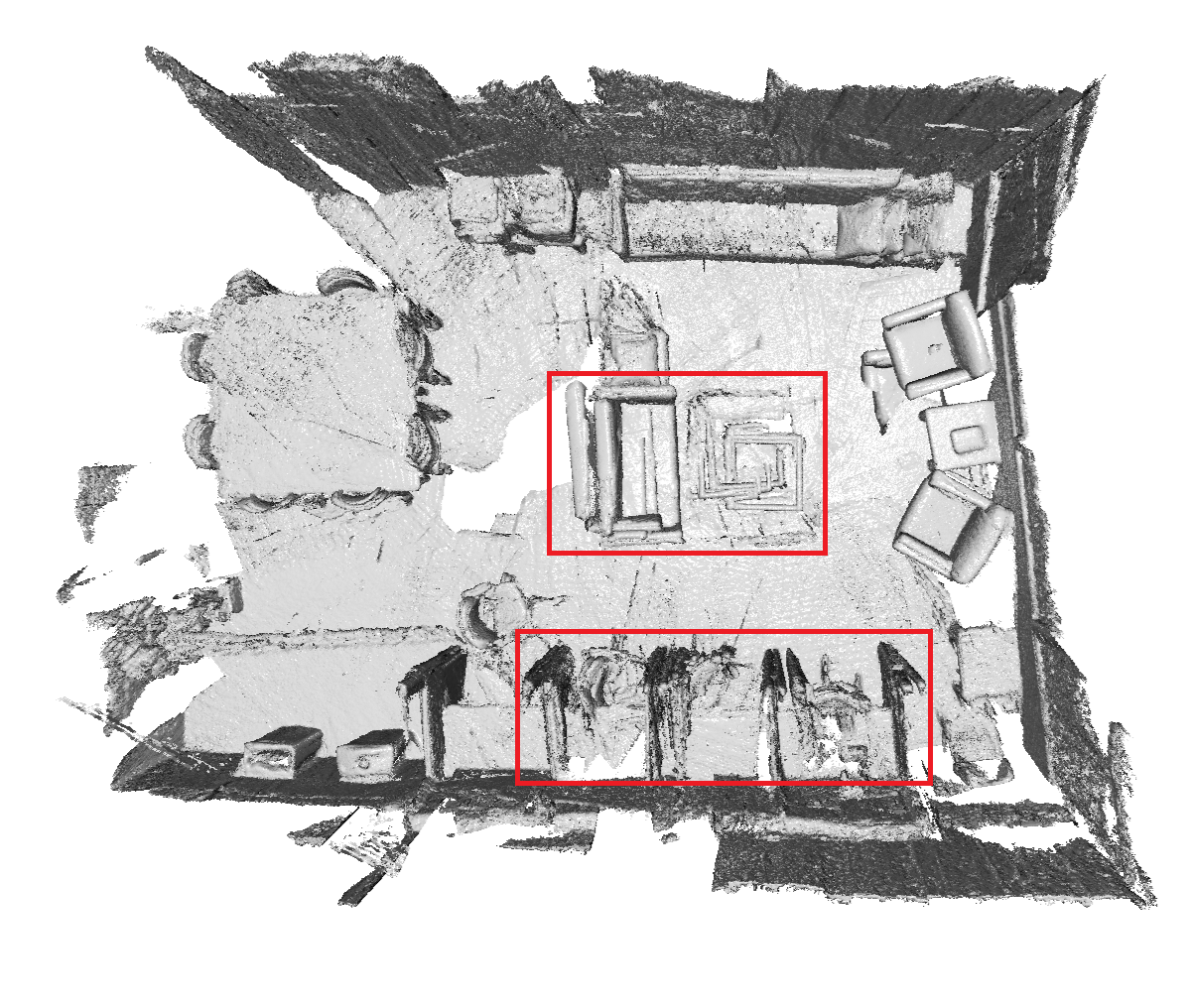}
    \includegraphics[width=1\linewidth]{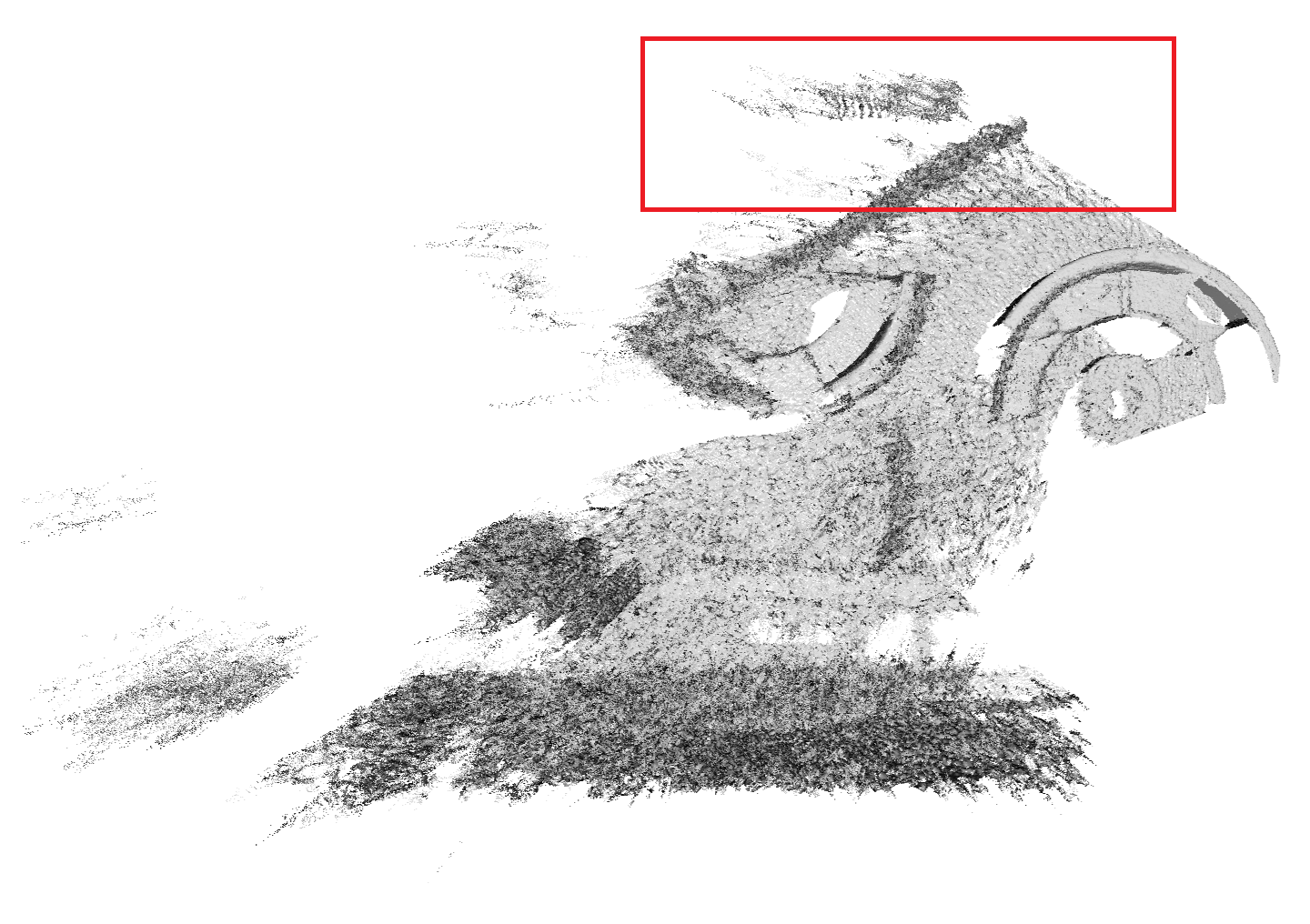}
    \end{minipage}}
    \subfigure[Ours]{
    \begin{minipage}[b]{0.23\linewidth}
    \includegraphics[width=1\linewidth]{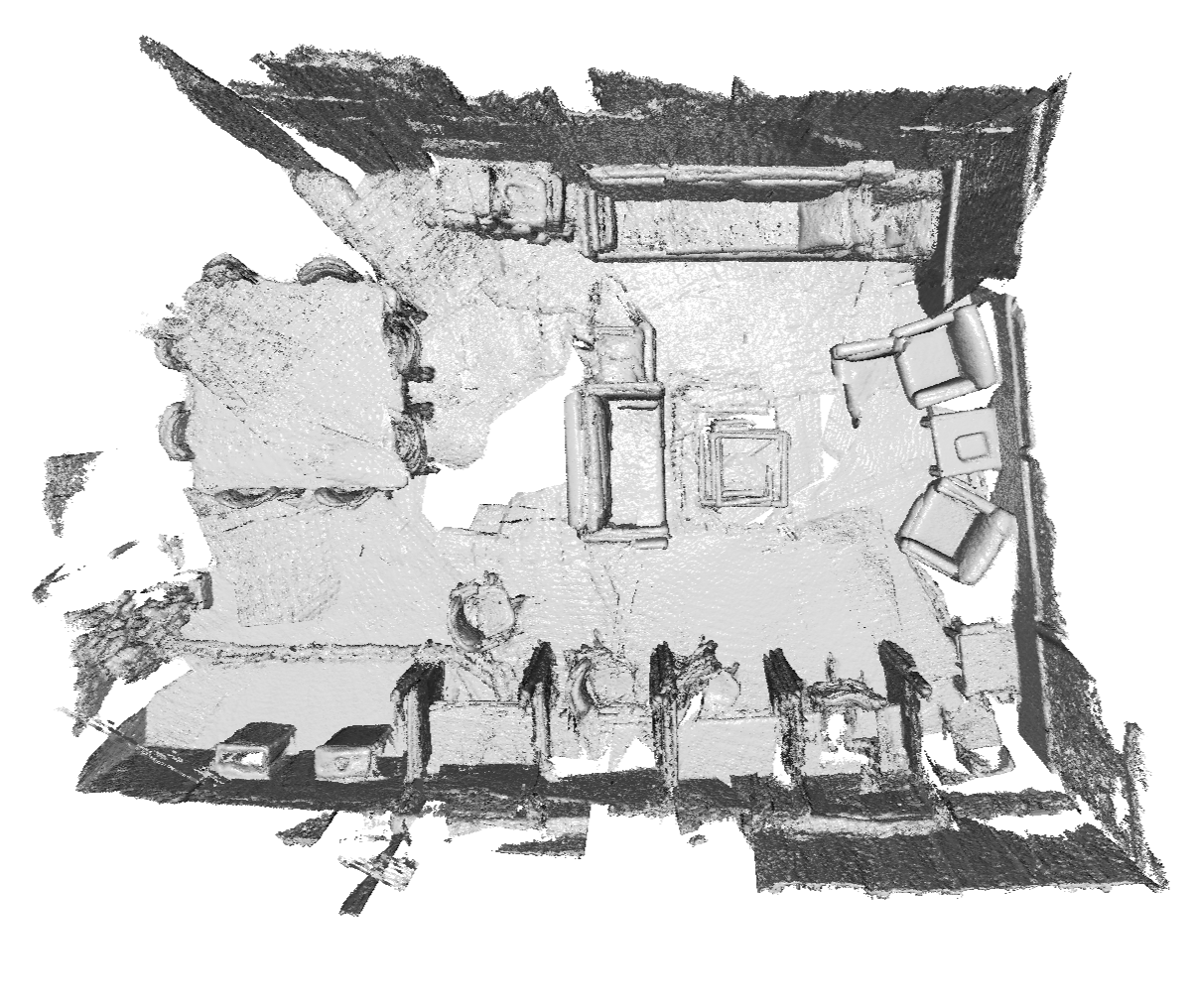}
    \includegraphics[width=1\linewidth]{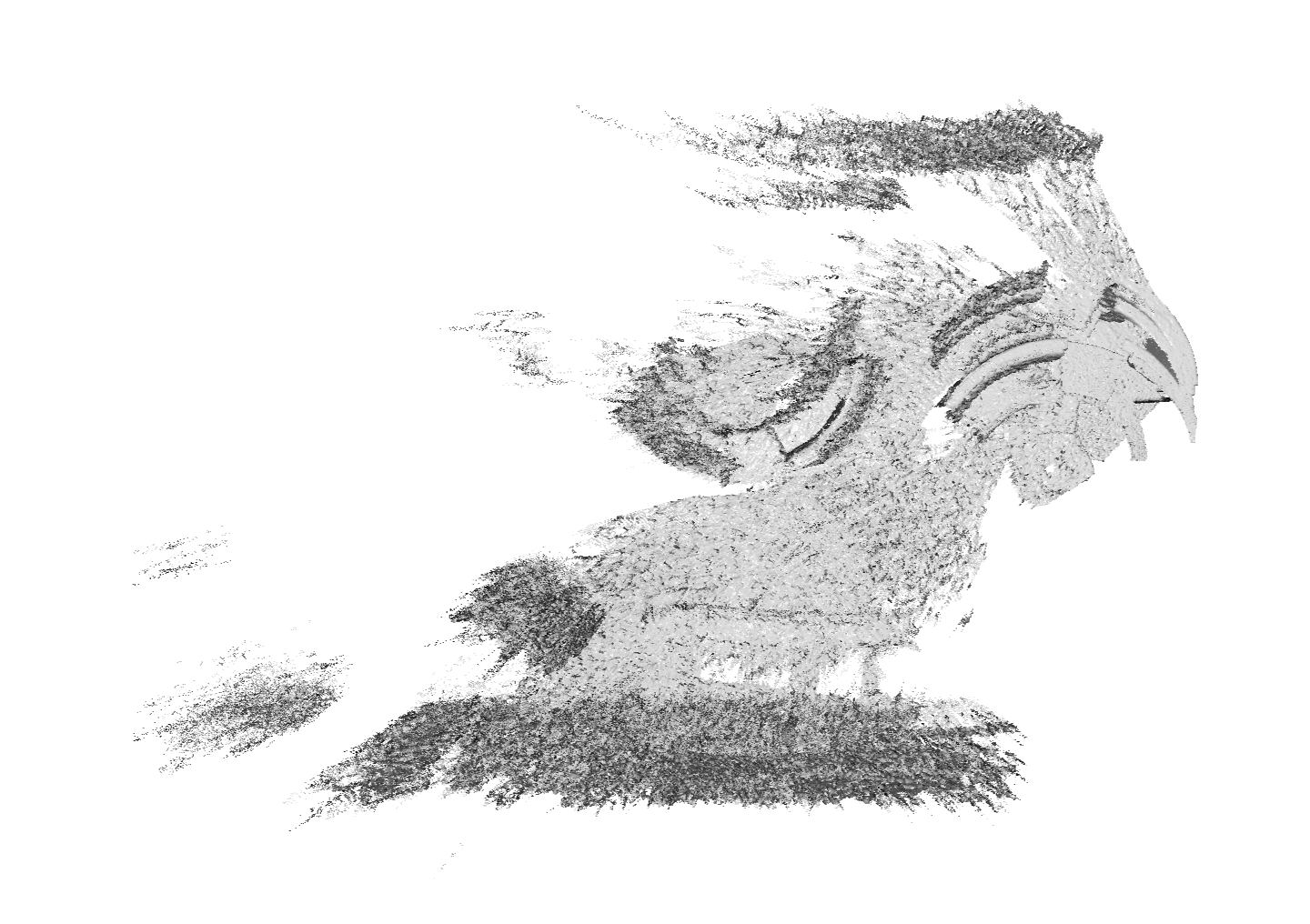}
    \end{minipage}}
    \subfigure[Ground Truth]{
    \begin{minipage}[b]{0.23\linewidth}
    \includegraphics[width=1\linewidth]{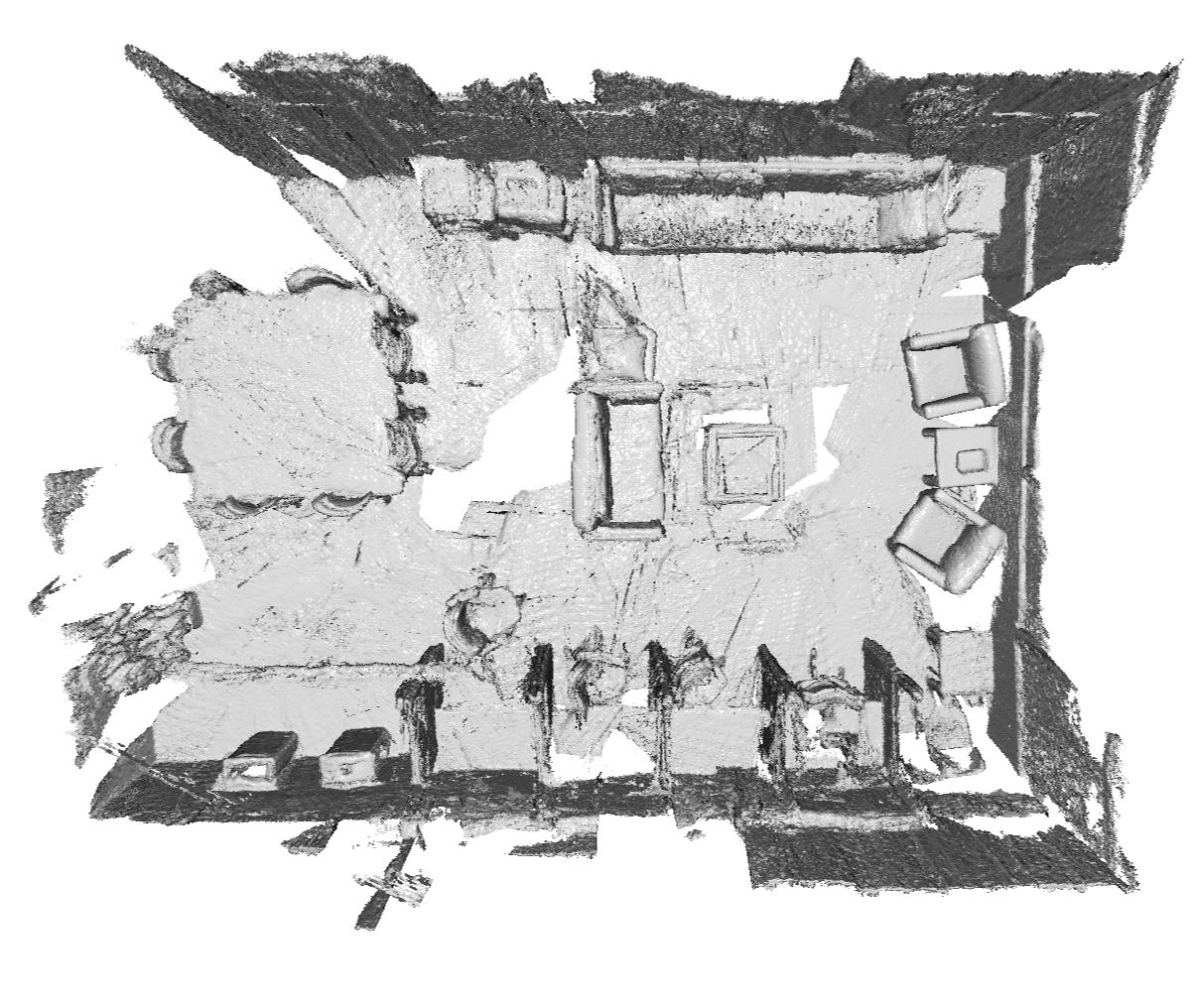}
    \includegraphics[width=1\linewidth]{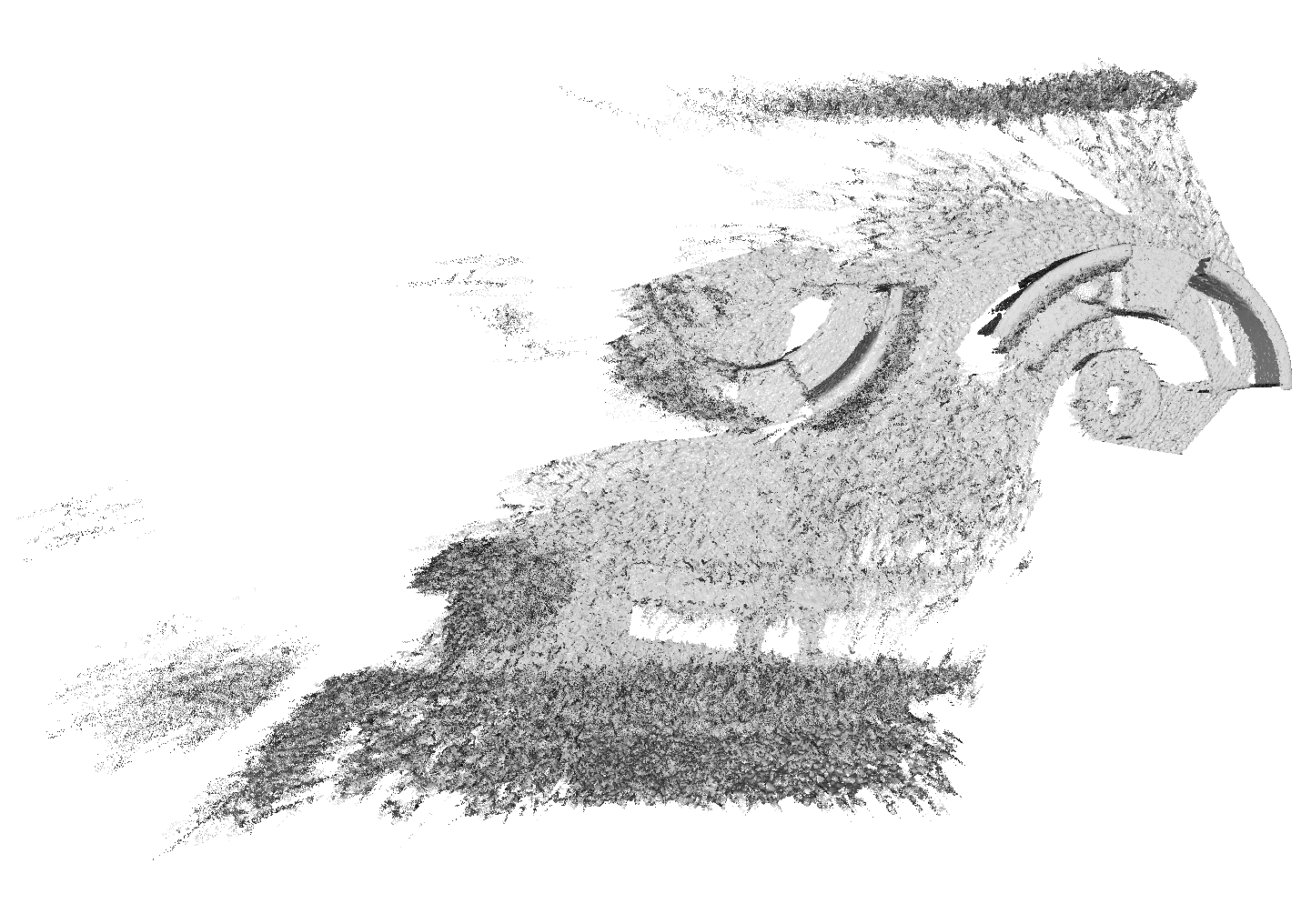}
    \end{minipage}}
    \caption{Qualitative comparison results. Top: 3DMatch \emph{studyroom2}. Bottom: ScanNet \emph{scene0334\_02}.}
    \label{fig:vis}
\end{figure*}

\subsection{Results}
The registration results on three datasets are presented in Table \ref{tab:3dmatch} and Table \ref{tab:scannet}. Our method consistently demonstrates outstanding performance across all three benchmarks. Notably, despite being trained solely on the 3DMatch dataset, our model exhibits strong performance in both indoor and outdoor scenarios, showcasing the remarkable effectiveness and generalization capability of our approach. On ScanNet, our model performs better at larger eCDF thresholds, likely due to the well-known difficulty deep networks have in regressing precise quantities. Nonetheless, our method achieves the best overall performance. Fig. \ref{fig:vis} provides some qualitative comparison results.

\subsection{Ablation Study}
To demonstrate the effectiveness of each component, we conduct ablation studies on 3DMatch and ScanNet. Table \ref{tab:ablation} provides a summary of the ablation results. Initially, we evaluate the impact of two geometric distribution features in experiments (a) and (b), both of which demonstrate positive influences, especially the ICR. Subsequently, in experiment (c), we assess the effectiveness of least squares refinement. Experiment (d) involves replacing the proposed overlap estimation module with NetVLAD in~\cite{wang2023robust}, revealing that our distance matching-based module has better performance and generalizability. In experiment (e), we replace our data-driven motion synchronization module with the history reweighting IRLS from~\cite{wang2023robust}. While slightly worse on 3DLoMatch, our design achieves comparable or better performance on 3DMatch and ScanNet. Additionally, we also attempted to initialize all absolute poses with the identity matrix~\cite{yew2021learning} or remove the GRU module, however, both settings failed to converge during training. 

\begin{table}[ht]
    \caption{Ablation study.}
    \label{tab:ablation}
    \centering
    \begin{tabular}{c|cccc}
    \hline
    \multirow{2}{*}{Exp}  & 3DMatch & 3DLoMatch & \multicolumn{2}{c}{ScanNet}\\
                          & RR    & RR      & $re$ Mean & $te$ Mean\\ 
    \hline
    
    (a) w/o IPR     & 95.6 & 81.0 & 17.5\degree & 0.38 \\
    (b) w/o ICR     & 94.1 & 77.9 & 17.6\degree & 0.40 \\
    (c) w/o refine  & 95.8 & 80.3 & 17.5\degree & 0.42 \\
    (d) w/ NetVLAD  & 94.8 & 81.2 & 19.1\degree & 0.41 \\
    (e) w/ HR-IRLS  & 96.2 & 83.5 & 23.2\degree & 0.59 \\
    Full            & 96.2 & 81.2 & 17.4\degree & 0.37 \\        
    \hline
    \end{tabular}
\end{table}

\subsection{Efficiency Test}
We conduct further analysis of the runtime to illustrate the efficiency of the proposed method. Table \ref{tab:time} summarizes the average time of multiview registration on 8 scenarios in the 3DMatch test set. Runtime measurements were taken on a machine equipped with an Intel(R) Core(TM) i5-9500 CPU @ 3.00GHz and NVIDIA GeForce GTX 1080 Ti. Despite our matching distance-based overlap estimation module being slower than the global feature method in SGHR~\cite{wang2023robust}, the computed correspondence can accelerate further pairwise registration. Additionally, our data-driven motion synchronization module exhibits significantly faster speed than the IRLS-based method. Therefore, our approach achieves comprehensive optimization.
\begin{table}[h]
    \caption{Time consumption on 3DMatch test set.}
    \label{tab:time}
    \centering
    \begin{tabular}{c|cccc}
    \hline
    Method & \begin{tabular}[c]{@{}c@{}}Overlap\\ Estimation\end{tabular} & \begin{tabular}[c]{@{}c@{}}Pairwise\\ Registration\end{tabular} & \begin{tabular}[c]{@{}c@{}}Motion\\ Synchronization\end{tabular} & Total \\
    \hline
    SGHR~\cite{wang2023robust}   & 5.06s         & 13.48s             & 4.63s                 & 23.17s \\
    Ours   & 9.09s         & 8.54s              & 0.03s                 & 17.66s \\
    \hline
    \end{tabular}
\end{table}

\section{Conclusion}
This paper presents a novel method for multiview point cloud registration. Our approach starts by constructing a sparse yet reliable pose graph utilizing descriptor matching distance-involved patterns between point cloud frames. Subsequently, we employ a data-driven motion synchronization module to alternately update rotation and translation features. This module intricately considers the geometric distribution within each pair and facilitates a flexible feature interaction by incorporating a confidence attention mechanism. Experimental results across diverse datasets confirm the effectiveness, generalization, and efficiency of our method.

\bibliographystyle{IEEEtran}
\bibliography{ref.bib}

\end{document}